# Development of a Large-scale Dataset of Chest Computed Tomography Reports in Japanese and a High-performance Finding Classification Model

Yosuke Yamagishi[1], Yuta Nakamura[2], Tomohiro Kikuchi[2,3], Yuki Sonoda[1], Hiroshi Hirakawa[1], Shintaro Kano[4], Satoshi Nakamura[4], Shouhei Hanaoka[1,5], Takeharu Yoshikawa[2], Osamu Abe[1,5]

1. Division of Radiology and Biomedical Engineering, Graduate School of Medicine, The University of Tokyo, Tokyo, Japan
2. Department of Computational Diagnostic Radiology and Preventive Medicine, The University of Tokyo Hospital, Tokyo, Japan
3. Department of Radiology, School of Medicine, Jichi Medical University, Shimotsuke, Tochigi, Japan
4. Department of Diagnostic Radiology, Toranomon Hospital, Tokyo, Japan
5. Department of Radiology, The University of Tokyo Hospital, Tokyo, Japan

Email: yamagishi-yosuke0115@g.ecc.u-tokyo.ac.jp

**Abstract**
**Background:** Recent advances in large language models have highlighted the need for high-quality multilingual medical datasets. While Japan leads globally in Computed Tomography (CT) scanner deployment and utilization, the absence of large-scale Japanese radiology datasets has hindered the development of specialized language models for medical imaging analysis. Despite the emergence of multilingual models and language-specific adaptations, the development of Japanese-specific medical language models has been constrained by the lack of comprehensive datasets, particularly in radiology.
**Objective:** To develop a comprehensive Japanese CT report dataset through machine translation and establish a specialized language model for structured finding classification, addressing the critical gap in Japanese medical natural language processing resources. Additionally, to create a rigorously validated evaluation dataset through expert radiologist review, ensuring reliable assessment of model performance.
**Methods:** We translated the CT-RATE dataset (24,283 CT reports from 21,304 patients) into Japanese using GPT-4o mini. The training dataset consisted of 22,778 machine-translated reports, while the validation dataset included 150 reports that were carefully revised by radiologists. We developed CT-BERT-JPN (Japanese), a specialized BERT model for extracting 18 structured findings from Japanese radiology reports, based on the "tohoku-nlp/bert-base-japanese-v3" architecture. Translated radiology reports were assessed using BLEU and ROUGE scores, complemented by expert radiologist review. Model performance was evaluated using standard metrics including accuracy, precision, recall, F1 score, and Area Under the Curve - Receiver Operating Characteristic, with GPT-4o serving as a baseline.
**Results:** Translation metrics showed preservation of general text structure, with BLEU scores of 0.731 and 0.690, and ROUGE scores ranging from 0.770 to 0.876 for Findings and from 0.748 to 0.857 for Impression sections, while expert review revealed necessary refinements in medical terminology. These modifications fell into three categories: contextual refinement of technical terms, completion of incomplete translations, and Japanese localization of medical terminology, highlighting the importance of expert validation in medical translation. CT-BERT-JPN demonstrated superior performance compared to GPT-4o in 11 out of 18 conditions, including lymphadenopathy (+14.2%), interlobular septal thickening (+10.9%), and atelectasis (+7.4%). The model achieved perfect scores across all metrics in four conditions (cardiomegaly, hiatal hernia, atelectasis, and interlobular septal thickening) and maintained F1 score exceeding 0.95 in 14 out of 18 conditions. Performance remained robust despite varying numbers of positive samples across conditions (ranging from 7 to 82 cases).

**Conclusions:** Our study establishes a robust Japanese CT report dataset and demonstrates the effectiveness of a specialized language model for structured finding classification. The hybrid approach of machine translation and expert validation enables the creation of large-scale medical datasets while maintaining high quality standards. This work provides essential resources for advancing medical AI research in Japanese healthcare settings, with both the dataset and model publicly available for research purposes to facilitate further advancement in the field.

## Introduction

Recent advances in large language models (LLMs) have demonstrated remarkable capabilities across various domains [1], leading to increased focus on developing multilingual models to serve diverse linguistic communities [2] . This trend is exemplified by the release of specialized language models such as Gemma-2-JPN, a Japanese-specific variant of Google's open LLM Gemma [3,4]. However, the development of such specialized models critically depends on the availability of high-quality, domain-specific datasets in target languages. This requirement becomes particularly crucial in specialized fields like medical imaging [5–7], where the interpretation of diagnostic findings demands both technical precision and linguistic accuracy.

Computed tomography (CT) scans play an indispensable role in modern medical diagnostics, facilitating disease staging, lesion evaluation, and early detection. Japan leads the world with the highest number of CT scanners per capita and an annual scan volume that surpasses most developed nations, presenting a vast reservoir of medical imaging data [8,9]. This extensive utilization of CT technology positions Japan as a pivotal contributor to global medical imaging resources. However, despite the proliferation of multilingual models and the growing emphasis on language-specific adaptations, there remains a notable absence of large-scale Japanese radiology report datasets [10]—a critical gap that hinders the development of Japanese-specific medical language models.

To address this challenge, we have constructed "CT-RATE-JPN," a Japanese version of the extensive "CT-RATE" dataset [11], which consists of CT scans and interpretation reports collected from 21,304 patients in Turkey. While general academic knowledge benchmarks have been successfully adapted for Japanese, as evidenced by JMMLU [12] and JMMMU [13] (Japanese versions of MMLU [14,15] and MMMU [16], respectively), and medical benchmarks like JMedBench [17] have emerged through combinations of translated English resources and Japanese medical datasets, a large-scale Japanese dataset specifically focused on radiology reports has remained notably absent.

CT-RATE-JPN employs an innovative approach to dataset construction, leveraging LLM-based machine translation to efficiently generate a large volume of training data. This addresses the fundamental challenge of dataset scale in medical AI development while maintaining quality through a strategic validation approach: a subset of the data undergoes careful revision by radiologists, creating a rigorously verified validation dataset. This dual-track methodology—combining machine-translated training data with specialist-validated evaluation data—establishes a robust pipeline for both training data acquisition and performance evaluation.

Both CT-RATE and CT-RATE-JPN retain licenses allowing free use for research purposes, supporting broader research initiatives in medical imaging and language processing. To demonstrate the practical utility of the CT-RATE-JPN dataset, we developed CT-BERT-JPN, a deep learning-based language model specifically designed to extract structured labels from Japanese radiology reports. By converting unstructured Japanese medical text into standardized, language-agnostic structured labels, CT-BERT-JPN provides a scalable framework for integrating Japanese radiology data into global medical AI development, addressing critical needs in the rapidly evolving landscape of multilingual medical AI.

## Methods

**Dataset Overview**

CT-RATE is a comprehensive dataset comprising 25,692 non-contrast chest CT volumes from 21,304 unique patients at Istanbul Medipol University Mega Hospital [11]. We selected this dataset as it is uniquely positioned as the only publicly available large-scale dataset that pairs CT volumes with radiology reports while permitting redistribution of derivative works. This dataset includes corresponding radiology text reports (consisting of a detailed findings section documenting observations and a concise impression section summarizing key information), multi-abnormality labels, and metadata. The dataset is divided into two cohorts: 20,000 patients for the training set and 1,304 patients for the validation dataset, allowing for robust model training and evaluation across diverse patient cases [18,19]. The training dataset, comprising 22,778 unique reports, was utilized for the construction of CT-RATE-JPN, the Japanese-translated version of the dataset, which was created using machine translation. For the validation dataset (n = 150), a randomly selected subset underwent machine translation, followed by a labor-intensive process of manual revision and refinement conducted by radiologists. The data selection process is described in Figure 1.

The CT-RATE dataset is annotated with 18 structured labels, covering key findings relevant to chest CT analysis. These labels include *Medical material*, *Arterial wall calcification*, *Cardiomegaly*, *Pericardial effusion*, *Coronary artery wall calcification*, *Hiatal hernia*, *Lymphadenopathy*, *Emphysema*, *Atelectasis*, *Lung nodule*, *Lung opacity*, *Pulmonary fibrotic sequela*, *Pleural effusion*, *Mosaic attenuation pattern*, *Peribronchial*

*thickening*, *Consolidation*, *Bronchiectasis*, and *Interlobular septal thickening*. The creators of the CT-RATE dataset developed a structured findings model based on the RadBERT architecture [20,21], trained on the manually labeled subset to label the remaining cases. This model achieved an F1 score ranging from 0.95 to 1.00, demonstrating its efficacy in accurately structuring these radiological findings from CT reports. This approach underscores the reliability of CT-RATE's structured annotations for high-performance diagnostic model development.

We also utilized these structured labels in the development of a Japanese structured findings model for CT-RATE-JPN, enabling accurate structuring of radiological findings in Japanese CT reports.

Given that CT-RATE is a publicly available dataset with de-identified patient information, and our study focused on the translation and linguistic analysis of the existing dataset without accessing any additional patient data, institutional review board approval was not required for this research.

**Figure 1.** Overview of the data selection process for CT-RATE-JPN. The figure illustrates the workflow for selecting cases from the CT-RATE dataset, including training and validation cohorts.

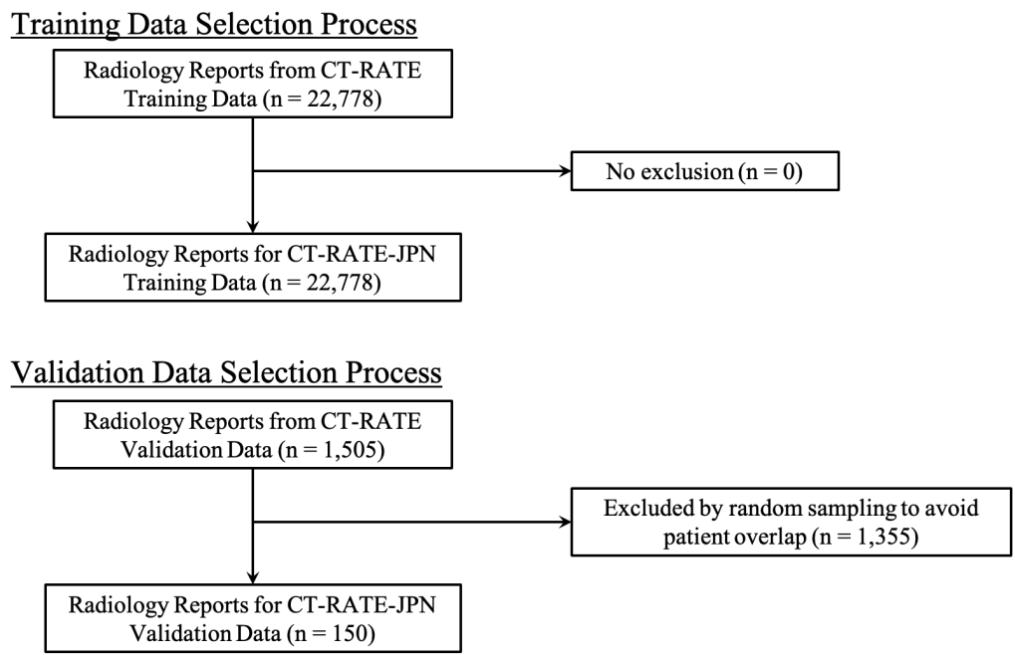

**Translation for CT-RATE-JPN**

For CT-RATE-JPN, we applied machine translation using GPT-4o mini (API version, "gpt-4o-mini-2024-07-18") [22], a lightweight, fast version of OpenAI's GPT-4o model [23]. GPT-4o mini is known for producing high-accuracy translations at an affordable rate, making it suitable for large-scale dataset translation. Each radiology report was processed by separately translating its findings and short impression sections. The complete translation prompts used for GPT-4o mini are provided in Supplementary

Figure 1 (Japanese original prompt) and Supplementary Figure 2 (English translation). The overall workflow is summarized in Figure 2.

To ensure accuracy and reliability in the evaluation data, we conducted a comprehensive manual correction process for 150 reports from the validation dataset. This process consisted of two distinct phases. In the first phase, we assembled a team of five radiology residents, all between their fourth and sixth post-graduate years, to conduct the initial review and revision of the machine translations. The reports were systematically distributed among team members to optimize workflow efficiency. We intentionally chose a larger team size to incorporate diverse clinical perspectives and minimize potential translation bias during the review process. The second phase involved a thorough expert review by two board-certified radiologists with extensive experience (post-graduate years 10 and 11). These senior radiologists divided the revised translations between them for final confirmation and refinement. The structured approach to task allocation, combined with this rigorous two-step review process, ensures that the validation dataset for CT-RATE-JPN meets the high-quality standards necessary for robust model assessment.

For the training dataset, all translations were generated solely using GPT-4o mini without manual corrections. This dataset is specifically designed to be used for machine learning model training. The decision to rely exclusively on machine-translated data for the training set balances the need for scale and the practical constraints of manual annotation.

Both CT-RATE and CT-RATE-JPN are released under a Creative Commons Attribution (CC BY-NC-SA) license, allowing free use for non-commercial research purposes with proper citation and shared terms for derivative works.

**Figure 2.** Workflow for the translation and validation process in constructing CT-RATE-JPN. The figure outlines the application of machine translation using GPT-4o mini for the training dataset and the two-phase manual correction process for 150 validation reports. Phase 1 involved initial revisions by radiology residents, while Phase 2 consisted of expert review and refinement by board-certified radiologists.

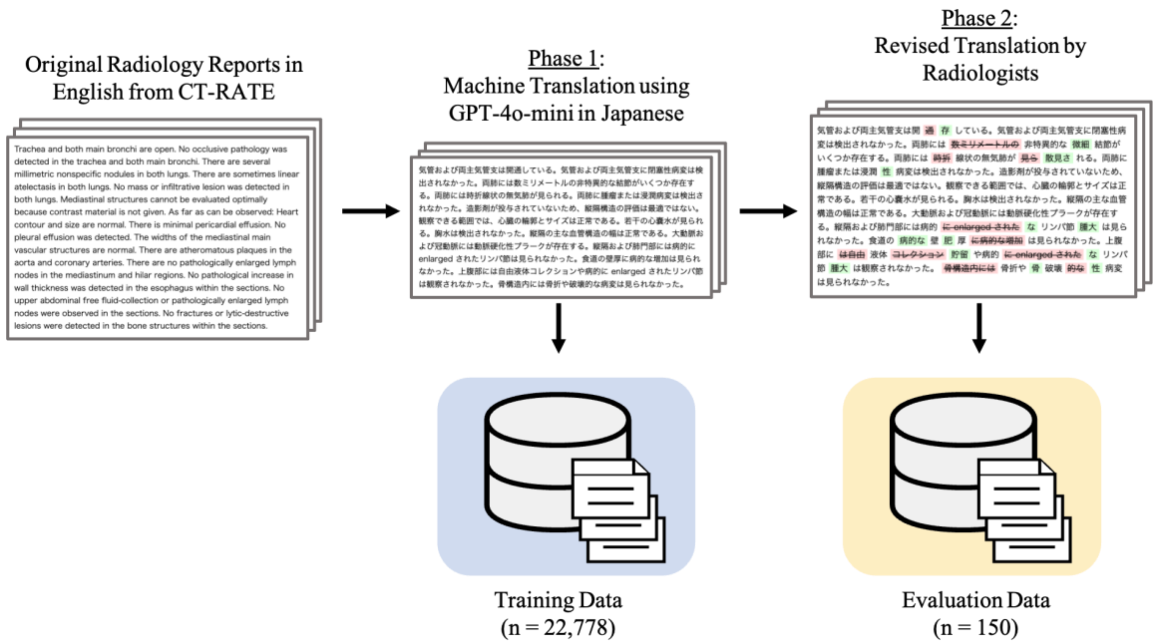

### Development of CT-BERT-JPN for Structured Finding Classification

For model training, we randomly split the training dataset into a 4:1 ratio, designating 80% for training and 20% for internal evaluation. We obtained the pretrained "tohoku-nlp/bert-base-japanese-v3" model from Hugging Face [24]. This model follows the architecture of the original BERT base model with 12 layers, 768 hidden dimensions, and 12 attention heads. It was pretrained on extensive Japanese datasets, including the Japanese portion of the CC-100 corpus [25,26] and the Japanese version of Wikipedia [27]. Notably, BERT-based models have demonstrated significant success in downstream tasks in the medical domain [28,29], making them a promising choice for our research.

Training was conducted using the transformers library (version 4.46.2) with a learning rate of 2e-5, a batch size of 8 for both training and evaluation, and a weight decay of 0.01. Binary Cross Entropy loss was applied to optimize the model for multi-label classification. The model was trained over four epochs, with internal evaluation and checkpoint saving at each epoch. The best-performing model on the internal evaluation data was selected and subsequently used for testing, which was conducted using the validation dataset of CT-RATE-JPN to ensure reliable performance assessment. The overall workflow for developing CT-BERT-JPN is illustrated in Figure 3.

**Figure 3.** Overview of the CT-BERT-JPN development workflow. The figure outlines the input data preparation from CT-RATE-JPN and the output as a fine-tuned BERT model (CT-BERT-JPN) for structured finding classification.

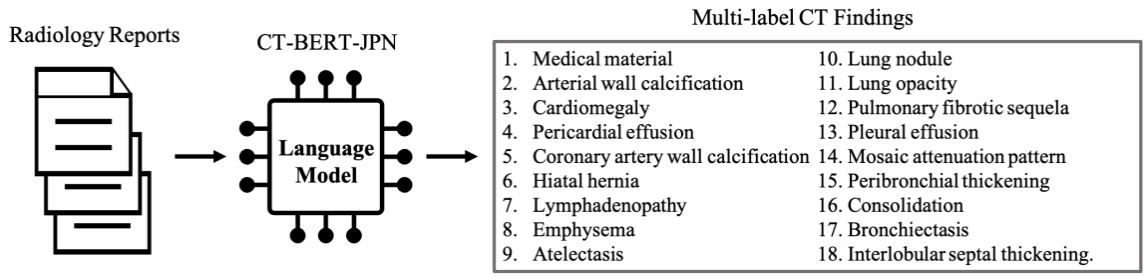

## Translated Radiology Reports Evaluation

For basic text analysis, we examined the structural characteristics of the translated reports, including character count, word count, sentence count, and lexical diversity. Since Japanese text, unlike English, does not use spaces to delimit words, we employed MeCab (version 1.0.10) [30], one of the most widely used morphological analyzers for Japanese text processing, to accurately segment and count words. These metrics were calculated for both machine-translated and radiologist-refined texts to assess the consistency of textual characteristics across different stages of dataset creation.

For translation quality assessment, we computed Bilingual Evaluation Understudy (BLEU) [31] and Recall-Oriented Understudy for Gisting Evaluation (ROUGE)-1, ROUGE-2, and ROUGE-L scores [32] using the nltk (version 3.9.1) and rouge-score (version 0.1.2) libraries.

BLEU is a metric that evaluates the accuracy of machine-translated text by comparing it to a reference translation. It measures how many words and short phrases in the machine translation match those in the reference translation, assessing the degree of similarity in wording and phrasing between the two texts.

ROUGE is a set of metrics used to assess the quality of summaries or translations by measuring the overlap between the machine-generated text and the reference text. ROUGE-1 considers the overlap of individual words, ROUGE-2 examines the overlap of pairs of consecutive words, and ROUGE-L focuses on the longest matching sequence of words between the two texts. These metrics emphasize recall, evaluating how much of the important content from the reference text is captured in the machine-generated text.

These metrics were calculated by comparing the machine-translated texts against radiologist-revised reference translations in the validation dataset.

## CT-BERT-JPN Performance Evaluation

For evaluating classification model performance for CT findings extraction, we utilized a test dataset comprising 150 radiology reports that had been revised by radiologists to ensure accuracy. Key metrics calculated for model assessment included accuracy, precision, recall, F1 score, and Area Under the Curve - Receiver Operating Characteristic (AUC-ROC). To establish a baseline, we performed structured labeling using GPT-4o (API version, "gpt-4o-2024-11-20"), which was selected due to its widespread adoption in various radiology tasks and its status as a representative closed-source commercial LLM. The input prompts used for GPT-4o are presented in Supplementary Figure 3 (original Japanese version) and Supplementary Figure 4 (English translation). These analyses were carried out using the scikit-learn (version 1.5.2) library.

# Results

**Dataset Overview**

The basic text statistics of the translated reports are summarized in Table 1, analyzing both Findings and Impression sections separately. The training dataset (n = 22,778) and validation dataset (consisting of 150 machine-translated reports and 150 radiologist-refined reports) showed consistent text structure across all metrics.

The Findings sections had character counts averaging around 455.6-475.0 characters, with slightly lower counts in radiologist-refined texts compared to machine translations. Word counts followed a similar pattern, averaging approximately 300 words per report across all datasets. The Impression sections were notably more concise, as expected for summary statements. Character counts averaged around 89.1-101.3 characters, with word counts of approximately 55.7-63.2 words per report. The sentence structure was also more condensed, with about 3.1-3.6 sentences per report.

Notably, in both sections, the overall text structure remained consistent between machine-translated and radiologist-refined versions, with similar patterns in sentence length and organization. While the refined versions showed slightly lower character and word counts compared to their machine-translated counterparts, the basic structural characteristics of the reports were preserved throughout the translation and refinement process.

**Table 1.** Basic text statistics of CT-RATE-JPN across different datasets, including both Findings and Impression sections. Values are presented as mean (standard deviation) per report. MT denotes machine-translated texts using GPT-4o mini, and Refined indicates texts after radiologist review and refinement. N represents the number of reports in each dataset.

| Section | Dataset | N | Characters | Words | Sentences | Unique Words |
|---|---|---|---|---|---|---|
| Findings | Training (MT) | 22,778 | 467.0 (148.0) | 303.3 (95.4) | 15.5 (4.6) | 126.8 (29.5) |
| Findings | Validation (MT) | 150 | 475.0 (130.1) | 307.0 (83.6) | 15.7 (3.9) | 128.9 (27.8) |
| Findings | Validation (Refined) | 150 | 455.6 (122.6) | 297.7 (80.8) | 15.7 (4.0) | 126.2 (27.4) |
| Impression | Training (MT) | 22,778 | 89.1 (68.9) | 55.7 (43.6) | 3.1 (2.2) | 38.1 (23.4) |
| Impression | Validation (MT) | 150 | 101.3 (76.0) | 63.2 (48.1) | 3.5 (2.6) | 41.7 (24.3) |
| Impression | Validation (Refined) | 150 | 97.8 (72.8) | 61.2 (46.0) | 3.6 (2.6) | 40.9 (24.0) |

The analysis of label distributions reveals a significant class imbalance in both training and validation dataset, as shown in Figure 4. In the training set, "Lung nodule" appears most frequently with 10,856 instances, while "Interlobular septal thickening" occurs least frequently with only 1,702 instances, representing a ratio of approximately 6.4:1. This imbalance is even more pronounced in the validation dataset, where the ratio between the most frequent (82 instances) and least frequent (7 instances) classes reaches about 11.7:1.

**Figure 4:** Bar plots showing the data distribution across different findings, sorted in descending order. Top: Training dataset distribution, Bottom: Validation dataset distribution. The number above each bar represents the number of positive samples for each condition.

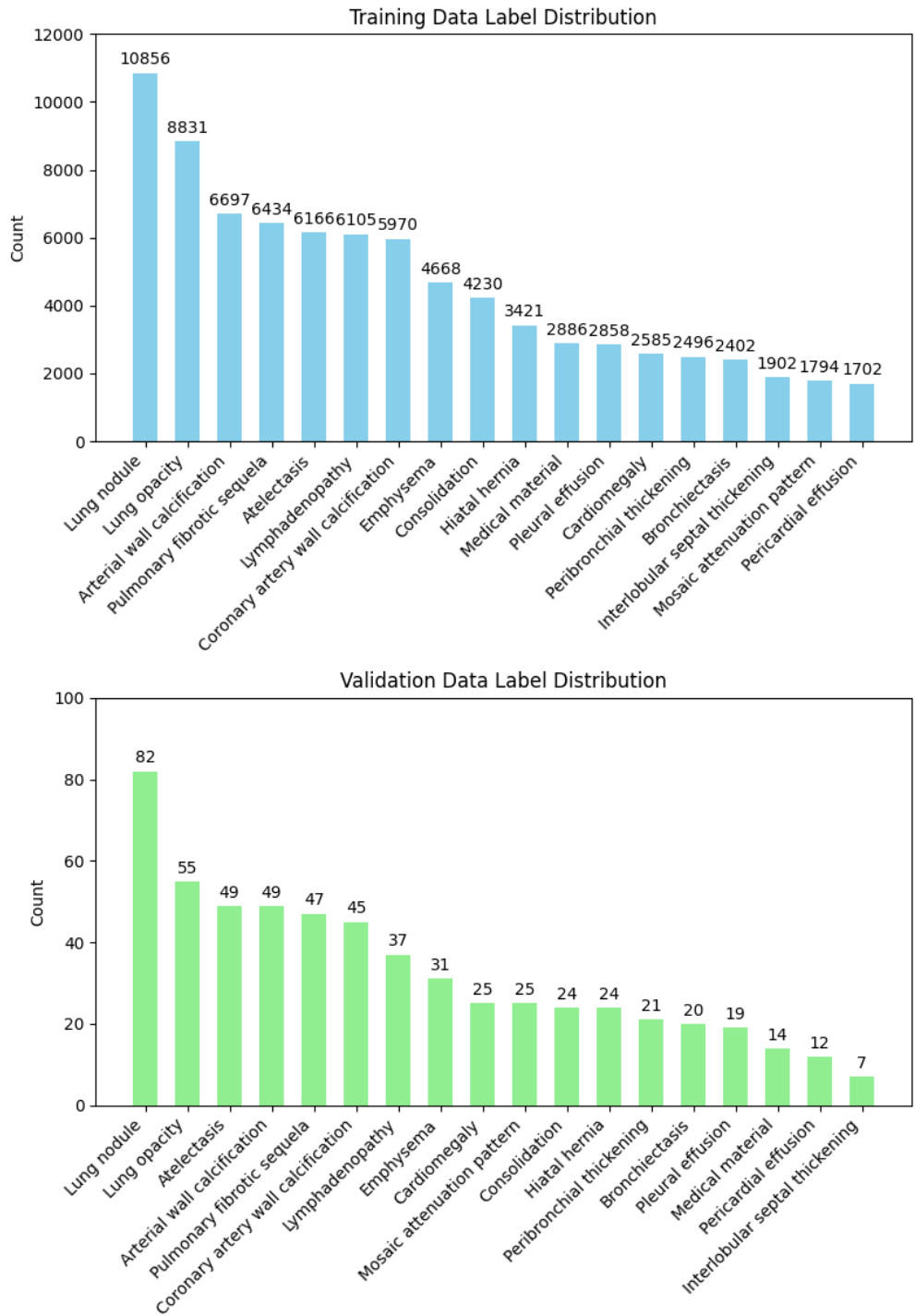

**Translated Radiology Reports Evaluation**

We evaluated the quality of machine-translated reports in CT-RATE-JPN through both automated metrics and expert assessment. For automated evaluation, we compared GPT-4o mini translations with radiologist-revised references in the validation dataset. The evaluation metrics for both sections are summarized in Table 2. These scores are all at high levels, demonstrating that the machine translation maintained the fundamental structure and meaning of the original reports.

**Table 2.** Automated evaluation metrics comparing GPT-4o mini translations with radiologist-revised references in the validation dataset. Values are presented as mean (standard deviation).

| Section | BLEU | ROUGE-1 | ROUGE-2 | ROUGE-L |
|---|---|---|---|---|
| Findings | 0.731 (0.104) | 0.876 (0.050) | 0.770 (0.091) | 0.854 (0.064) |
| Impression | 0.690 (0.196) | 0.857 (0.104) | 0.748 (0.161) | 0.837 (0.120) |

However, despite the high automated metric scores, qualitative analysis revealed that substantial revisions were necessary for medical terminology. The radiologists' assessment identified representative patterns of necessary modifications. This expert review revealed three major categories of improvements (Figure 5), which reflect typical challenges in medical translation: 1) contextual refinement of technically correct but unnatural terms, 2) completion of missing or incomplete translations, and 3) Japanese localization of untranslated English terms. While not exhaustive, these patterns represent key areas where human expertise complements machine translation in medical contexts.

The first category, contextual refinement, primarily involved replacing technically accurate but clinically uncommon expressions with more natural medical terminology. For instance, direct translations of anatomical conditions were often revised to their proper radiological equivalents when describing vessel status, reflecting standard terminology in Japanese clinical practice. The second category addressed cases where certain medical terms were either missing or incompletely translated, requiring additional context-specific information. A typical example would be where anatomical descriptions lacked specific diagnostic terminology common in radiological reporting. The third category focused on proper localization of English medical terms that were initially left untranslated, such as converting technical descriptors of pathological findings into their appropriate Japanese radiological counterparts.

**Figure 5.** Representative examples of translation modifications in CT-RATE-JPN. The figure illustrates three major categories of translation improvements identified during radiologist review: (1) contextual refinement of technically correct but unnatural terms, such as replacing literal translations with standard medical terminology commonly used in Japanese radiology reports; (2) completion of missing or incomplete translations, where critical diagnostic or anatomical details were added to ensure clarity and completeness; and (3) localization of untranslated English terms, involving the conversion of technical descriptors and pathological findings into their appropriate Japanese radiological equivalents.

Translated Report after Radiologist Revision

1. **Contextual refinement: technically correct but unnatural terms**
   気管および両主気管支は開 通 存 している。
   (Trachea and both main bronchi are open.)

2. **Translation completion: missing or incomplete translations**
   両肺に腫瘤または浸潤 性 病変は検出されなかった。
   (No mass or infiltrative lesion was detected in both lungs.)

3. **Japanese localization: untranslated English terms**
   病的 に enlarged された な リンパ節 腫大
   (pathologically enlarged lymph nodes)

**CT-BERT-JPN Performance Evaluation**

Table 3 presents the performance evaluation results of CT-BERT-JPN across 18 different findings from 150 chest CT radiology reports. The model achieved perfect scores (1.000) across all evaluation metrics (accuracy, precision, recall, F1 score, and AUC-ROC) in three findings: Pericardial effusion, Hiatal hernia, and Mosaic attenuation pattern. The model demonstrated high accuracy exceeding 0.95 in 17 out of 18 findings, with AUC-ROC values surpassing 0.98 in all findings. Within the dataset, the number of positive samples varied considerably across conditions, from 7 cases (interlobular septal thickening) to 82 cases (lung nodule). Despite this imbalanced distribution of positive samples, the model maintained robust performance metrics across all conditions.

**Table 3.** Performance evaluation of CT-BERT-JPN across 18 different findings. The table shows the model's performance metrics, including Accuracy, Precision, Recall, F1 score, and AUC-ROC.

| Findings | Accuracy | Precision | Recall | F1 | AUC-ROC |
|---|---|---|---|---|---|
| Medical material | 0.973 | 0.778 | 1.000 | 0.875 | 0.999 |
| Arterial wall calcification | 0.987 | 0.961 | 1.000 | 0.980 | 1.000 |
| Cardiomegaly | 0.987 | 1.000 | 0.920 | 0.958 | 0.999 |
| Pericardial effusion | 1.000 | 1.000 | 1.000 | 1.000 | 1.000 |
| Coronary artery wall calcification | 0.987 | 0.978 | 0.978 | 0.978 | 1.000 |
| Hiatal hernia | 1.000 | 1.000 | 1.000 | 1.000 | 1.000 |
| Lymphadenopathy | 0.987 | 0.973 | 0.973 | 0.973 | 0.994 |
| Emphysema | 0.980 | 0.938 | 0.968 | 0.952 | 0.989 |
| Atelectasis | 0.993 | 0.980 | 1.000 | 0.990 | 1.000 |
| Lung nodule | 0.967 | 0.975 | 0.963 | 0.969 | 0.991 |
| Lung opacity | 0.953 | 0.929 | 0.945 | 0.937 | 0.991 |
| Pulmonary fibrotic sequela | 0.953 | 0.935 | 0.915 | 0.925 | 0.981 |
| Pleural effusion | 0.987 | 0.905 | 1.000 | 0.950 | 1.000 |
| Mosaic attenuation pattern | 1.000 | 1.000 | 1.000 | 1.000 | 1.000 |
| Peribronchial thickening | 0.960 | 1.000 | 0.714 | 0.833 | 0.985 |
| Consolidation | 0.933 | 0.706 | 1.000 | 0.828 | 0.996 |
| Bronchiectasis | 0.980 | 0.870 | 1.000 | 0.930 | 0.990 |
| Interlobular septal thickening | 0.993 | 0.875 | 1.000 | 0.933 | 1.000 |

Figure 6 illustrates a bar chart comparing F1 scores between CT-BERT-JPN and GPT-4o across 18 conditions. CT-BERT-JPN showed higher F1 scores than GPT-4o in 11 findings and achieved equivalent performance in 2 findings, specifically Hiatal hernia and Mosaic attenuation pattern, both of which attained perfect F1 scores. Notably, CT-BERT-JPN showed higher performance in Lymphadenopathy (0.142 higher), Interlobular septal thickening (0.109 higher), and Atelectasis (0.074 higher). However, the model showed lower performance in 5 findings, with the largest differences observed in Peribronchial thickening (0.051 lower) and Consolidation (0.035 lower). Detailed performance metrics for GPT-4o are presented in Supplementary Table 1.

**Figure 6.** Bar chart comparing the F1 scores of CT-BERT-JPN and GPT-4o across 18 findings. The chart visualizes the performance of both models, with CT-BERT-JPN shown in blue and GPT-4o in orange, highlighting their respective scores for each finding.

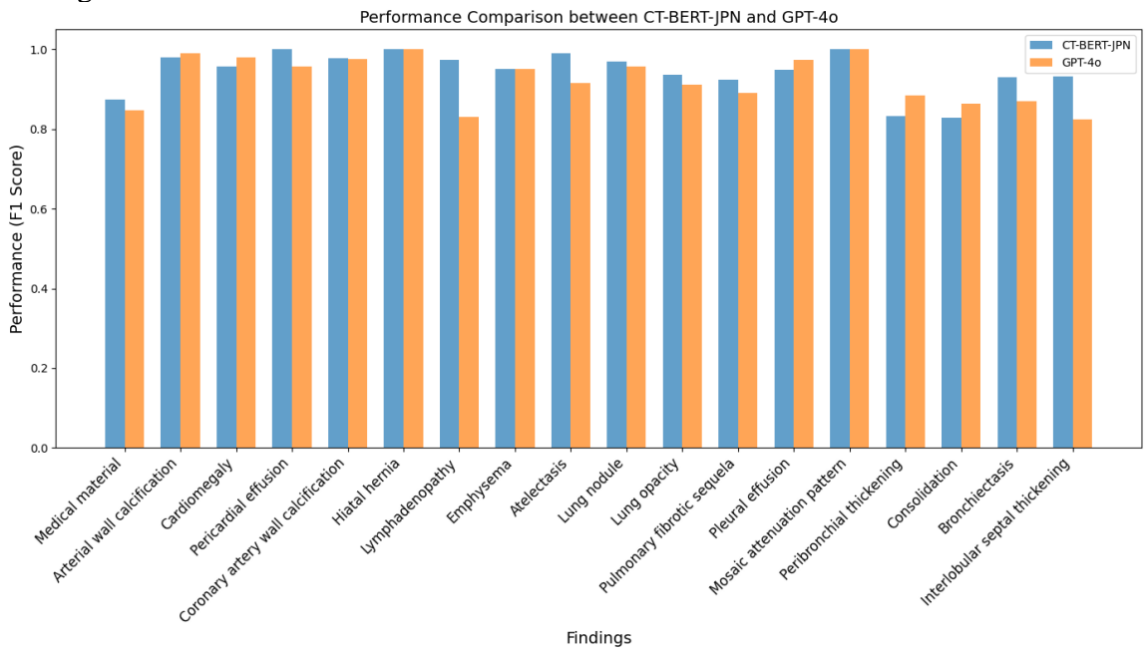

We conducted a detailed analysis of CT-BERT-JPN's performance comparing two scenarios: using radiologist-refined translated reports as input versus using raw machine-translated reports. The differences in metrics are detailed in Table 4 (performance metrics for machine translation can be found in Supplementary Table 2). The analysis shows that for most findings, performance differences were less than 5%, with no significant variations, confirming the model's robustness. However, Peribronchial thickening showed notable decreases with a recall drop of -0.238 and an F1 score reduction of -0.143, while Consolidation experienced relatively significant performance declines with precision dropping by -0.179 and F1 score by -0.092. Nevertheless, the AUC-ROC decreased by no more than -0.011 across all items, suggesting that model performance can be maintained even with radiologist-corrected reports as input through optimal threshold search.
.

**Table 4.** Performance metric differences of CT-BERT-JPN between models using radiologist-refined translations versus raw machine translations as input. Values represent the difference (radiologist-corrected minus machine-translated) in Accuracy, Precision, Recall, F1 score, and AUC-ROC for each finding. Positive values (+) indicate higher performance with radiologist-corrected translations, while negative values (-) indicate higher performance with raw machine translations.

| Findings | Accuracy | Precision | Recall | F1 | AUC-ROC |
|---|---|---|---|---|---|
| Medical material | 0.000 | -0.034 | +0.071 | +0.008 | +0.002 |
| Arterial wall calcification | -0.006 | -0.019 | 0.000 | -0.010 | 0.000 |
| Cardiomegaly | -0.013 | 0.000 | -0.080 | -0.042 | -0.001 |
| Pericardial effusion | +0.007 | 0.000 | +0.083 | +0.043 | 0.000 |
| Coronary artery wall calcification | 0.000 | 0.000 | 0.000 | 0.000 | 0.000 |
| Hiatal hernia | 0.000 | 0.000 | 0.000 | 0.000 | 0.000 |
| Lymphadenopathy | -0.006 | -0.001 | -0.027 | -0.014 | -0.006 |
| Emphysema | -0.007 | -0.030 | 0.000 | -0.016 | -0.011 |
| Atelectasis | -0.007 | -0.020 | 0.000 | -0.010 | 0.000 |
| Lung nodule | -0.013 | -0.025 | 0.000 | -0.012 | 0.000 |
| Lung opacity | -0.007 | -0.016 | 0.000 | -0.008 | -0.002 |
| Pulmonary fibrotic sequela | -0.007 | 0.017 | -0.042 | -0.013 | -0.005 |
| Pleural effusion | -0.006 | -0.045 | 0.000 | -0.024 | +0.002 |
| Mosaic attenuation pattern | 0.000 | 0.000 | 0.000 | 0.000 | 0.000 |
| Peribronchial thickening | -0.033 | 0.000 | -0.238 | -0.143 | -0.010 |
| Consolidation | -0.040 | -0.179 | +0.042 | -0.092 | +0.004 |
| Bronchiectasis | +0.007 | +0.037 | 0.000 | 0.021 | -0.006 |
| Interlobular septal thickening | -0.007 | -0.125 | 0.000 | -0.067 | 0.000 |

## Discussion

**Principal findings**

Our study yielded three key findings in developing Japanese medical imaging resources and analysis capabilities:

First, we established an efficient workflow combining machine translation and expert validation that successfully created a large-scale Japanese radiology dataset while maintaining high quality standards through focused radiologist review.

Second, our specialized CT-BERT-JPN model demonstrated superior performance to GPT-4o in most structured finding extraction tasks, despite its relatively compact architecture, highlighting the effectiveness of domain-specific optimization.

Third, the model maintained robust performance across both machine-translated and radiologist-revised reports, suggesting the viability of machine translation for training data creation in specialized medical domains.

**Dataset Development and Quality Assessment**

In the field of medical imaging datasets, several English-language resources have been previously established, including OpenI dataset [33] and MIMIC-CXR [34] for chest radiographs and AMOS-MM [35] for chest-to-pelvic CT scans; however, these datasets were exclusively available in English, with a limited multilingual adaptations available. While many LLMs and VLMs claim multilingual capabilities [26,36,37], their performance consistently degrades when handling non-English languages, including Japanese [38]. One significant factor contributing to this performance gap is the scarcity of non-English datasets, making the development of such resources an urgent priority.

Machine-translated reports showed minimal differences in basic structural elements compared to the original texts, with word counts and sentence lengths varying by less than 5%. The translation quality evaluation using BLEU and ROUGE metrics also demonstrated high performance, indicating that the overall structure and content of the reports were well preserved. However, qualitative analysis revealed that GPT-4o mini alone could not guarantee the quality of translated radiology reports, particularly in handling specialized medical terminology. This limitation necessitated our rigorous refinement process involving expert radiologists, which proved crucial in establishing a reliable validation dataset.

Our efficient workflow enabled the review and correction of approximately 70,000 characters of machine-translated text in total, significantly reducing the time and effort typically required for such extensive translation verification. This efficiency was achieved through a structured review process where radiologists could focus primarily on medical terminology and critical semantic elements, rather than reviewing every aspect of the translation. The systematic approach we employed—combining large language models with domain expert validation—presents a scalable methodology for dataset creation between distinctly different languages like English and Japanese. This hybrid process offers a promising framework that could be adapted not only for other languages but also for specialized domains beyond radiology where precise terminology and domain expertise are critical.

**Model Performance and Evaluation**

CT-BERT-JPN, when combined with CT volumes, demonstrates significant potential for various text-based applications in medical imaging analysis. Our comprehensive evaluation revealed exceptional performance in structured finding extraction across diverse radiological conditions, with F1 scores consistently above 0.95 and perfect scores in several findings. More notably, CT-BERT-JPN outperformed GPT-4o in 11 out of 18 findings, achieving higher F1 scores in significant conditions such as lymphadenopathy (+0.142), interlobular septal thickening (+0.109), and atelectasis (+0.074). This superior performance is particularly remarkable considering that CT-BERT-JPN has only approximately 110 million parameters, whereas recent large language models often employ hundreds of billions of parameters [39]. Our results demonstrate that a specialized, compact language model can achieve state-of-the-art performance in domain-specific tasks, even when compared to more sophisticated general-purpose models.

Importantly, the model maintained stable performance across both unmodified machine translations and radiologist-revised reports, with AUC-ROC decreases no greater than 0.011 across all findings. While some conditions showed moderate performance variations between machine-translated and radiologist-revised reports, particularly in findings where machine translation errors were common (F1 score differences of -0.143 for peribronchial thickening and -0.092 for consolidation), with consolidation being notably affected by mistranslations of the related term "infiltration", the overall robustness of the model suggests that effective clinical applications can be developed using machine-translated training data. Although established models such as RadGraph [40] and CheXpert labelling tool [41] have shown success with English reports, our model represents the first openly available Japanese model for multi-label CT findings extraction.

**Clinical Implementation and Impact**

The development of such a structured findings extraction model in Japan, where medical imaging utilization rates rank among the highest globally, contributes substantially to both domestic healthcare advancement and international model development. Our implementation incorporates previous work in Japanese radiology report processing, including end-to-end approaches for clinical information extraction [42], natural language processing systems for pulmonary nodule follow-up assessment [43], and BERT-based transfer learning methods for anatomical classification [44].

Our multi-label classification approach addresses several implementation challenges unique to the Japanese healthcare context. These include vocabulary standardization across institutions, integration with existing reporting systems, and the development of specialized Japanese medical vocabulary handling mechanisms. The model's ability to process Japanese-specific medical expressions and maintain high performance across different reporting styles demonstrates its potential for broad clinical application. However, prospective validation across multiple Japanese institutions remains essential for both performance evaluation and capturing institution-specific reporting patterns that inform targeted model improvements.

**Limitations and Future Directions**

Several limitations should be acknowledged in our study. First, while we demonstrated the effectiveness of our language model through rigorous evaluation, the utility of CT-RATE-JPN for vision-language models like CT-CLIP [18], which require joint learning of CT volumes and text descriptions, remains to be empirically validated. Second, our reliance on the CT-RATE dataset may introduce inherent biases in reporting styles and patterns, as radiology reports typically vary considerably across institutions and individual radiologists. Third, the translation-based approach, despite expert validation, may not fully capture the nuanced expressions and specialized terminology common in Japanese clinical practice.

These limitations suggest several promising directions for future work. A primary direction is the development of vision-language models using CT-RATE-JPN in conjunction with CT-RATE's CT volumes. Given the successful development of Japanese CLIP models for general domain tasks, which have demonstrated the feasibility of cross-lingual vision-language alignment in Japanese [45], extending this approach to medical imaging is particularly promising. While this endeavor requires substantial computational resources and sophisticated training strategies, various approaches can be explored, such as additional training on existing models like CT-CLIP. Furthermore, the construction of a new dataset comprising pairs of Japanese radiology reports and CT volumes from Japanese medical institutions would enable more direct assessment of model performance in the Japanese healthcare context and potentially reveal insights unique to this setting. Our benchmark dataset, validated by radiologists through a systematic review process, provides a valuable foundation for evaluating such Japanese-English and English-Japanese translation models in the radiology domain.

**Conclusions**

In this study, we introduced CT-RATE-JPN, a comprehensive Japanese dataset of CT interpretation reports, and developed a specialized language model for structured labeling. Our model demonstrated superior performance compared to GPT-4o, achieving higher F1 scores in numerous categories of structured finding extraction. The creation of CT-RATE-JPN, along with our publicly available structured findings model, represents a significant contribution to Japanese medical imaging research. By making both the dataset and model freely accessible to the research community, we enable reproducibility and foster collaborative advancement in the field. This work not only provides essential resources for the medical AI community but also establishes a robust foundation for developing more sophisticated multilingual medical vision-language models. These openly available contributions will support the development of AI-assisted diagnostic tools while maintaining the high standards required for clinical applications in radiology.

**Data Availability**

The original CT-RATE dataset, including CT volumes and structured data, is publicly available through Hugging Face Datasets [46]. Our Japanese translation dataset, CT RATE-JPN, is also publicly available through Hugging Face Datasets [47]. Both datasets are released under the CC BY-NC-SA license, allowing free use for non-commercial research purposes. Usage of the datasets requires proper citation and redistribution under similar license terms.

The trained CT-BERT-JPN model is also publicly available through Hugging Face Models for research purposes only [48]. The model can be freely used and adapted for academic and research applications with appropriate citation. Commercial use of the model is not permitted.

## Acknowledgements

The authors declare that no funding was used to support this research. GPT-4o by OpenAI was used for manuscript proofreading.

## Funding

This research did not receive any specific grant from funding agencies in the public, commercial, or not-for-profit sectors.

## Conflicts of Interest

The Department of Computational Diagnostic Radiology and Preventive Medicine, The University of Tokyo Hospital, is sponsored by HIMEDIC Inc and Siemens Healthcare K.K.

## Authors' Contributions

All authors agreed on the research conceptualization and design. Y. Y. conducted algorithm planning and implementation, model development, statistical analysis, data curation, and drafted the manuscript. Y. N. contributed to algorithm planning and data curation. T. K. performed statistical analysis and data curation. Y. S., H. H., S. K., and S. N. contributed to data curation. S. H., T. Y., and O. A. provided supervision and revised the manuscript. All authors reviewed and approved the final version of the manuscript.

## Abbreviations

AUC-ROC: Area Under the Curve - Receiver Operating Characteristic
BLEU: Bilingual Evaluation Understudy
CC BY-NC-SA: Creative Commons Attribution-NonCommercial-ShareAlike
CT: Computed Tomography
LLM: Large Language Model
ROUGE: Recall-Oriented Understudy for Gisting Evaluation
VLM: Vision Language Model

**Supplementary Materials**
**Supplementary Table 1.** Performance evaluation of GPT-4o across 18 different pathological findings and anatomical structures. The table shows the model's performance metrics, including Accuracy, Precision, Recall, and F1 score.

| Findings | Accuracy | Precision | Recall | F1 |
|---|---|---|---|---|
| Medical material | 0.967 | 0.737 | 1.000 | 0.848 |
| Arterial wall calcification | 0.993 | 0.980 | 1.000 | 0.990 |
| Cardiomegaly | 0.993 | 1.000 | 0.960 | 0.980 |
| Pericardial effusion | 0.993 | 1.000 | 0.917 | 0.957 |
| Coronary artery wall calcification | 0.987 | 1.000 | 0.956 | 0.977 |
| Hiatal hernia | 1.000 | 1.000 | 1.000 | 1.000 |
| Lymphadenopathy | 0.927 | 0.964 | 0.730 | 0.831 |
| Emphysema | 0.980 | 0.967 | 0.935 | 0.951 |
| Atelectasis | 0.940 | 0.845 | 1.000 | 0.916 |
| Lung nodule | 0.953 | 0.963 | 0.951 | 0.957 |
| Lung opacity | 0.933 | 0.895 | 0.927 | 0.911 |
| Pulmonary fibrotic sequela | 0.933 | 0.911 | 0.872 | 0.891 |
| Pleural effusion | 0.993 | 0.950 | 1.000 | 0.974 |
| Mosaic attenuation pattern | 1.000 | 1.000 | 1.000 | 1.000 |
| Peribronchial thickening | 0.967 | 0.864 | 0.905 | 0.884 |
| Consolidation | 0.953 | 0.815 | 0.917 | 0.863 |
| Bronchiectasis | 0.960 | 0.769 | 1.000 | 0.870 |
| Interlobular septal thickening | 0.980 | 0.700 | 1.000 | 0.824 |

**Supplementary Table 2.** Performance evaluation of CT-BERT-JPN using raw machine-translated reports as input across 18 different findings. The table shows the model's performance metrics, including Accuracy, Precision, Recall, F1 score, and AUC-ROC.

| Findings | Accuracy | Precision | Recall | F1 | AUC-ROC |
|---|---|---|---|---|---|
| Medical material | 0.973 | 0.812 | 0.929 | 0.867 | 0.997 |
| Arterial wall calcification | 0.993 | 0.980 | 1.000 | 0.990 | 1.000 |
| Cardiomegaly | 1.000 | 1.000 | 1.000 | 1.000 | 1.000 |
| Pericardial effusion | 0.993 | 1.000 | 0.917 | 0.957 | 1.000 |
| Coronary artery wall calcification | 0.987 | 0.978 | 0.978 | 0.978 | 1.000 |
| Hiatal hernia | 1.000 | 1.000 | 1.000 | 1.000 | 1.000 |
| Lymphadenopathy | 0.993 | 0.974 | 1.000 | 0.987 | 1.000 |
| Emphysema | 0.987 | 0.968 | 0.968 | 0.968 | 1.000 |
| Atelectasis | 1.000 | 1.000 | 1.000 | 1.000 | 1.000 |
| Lung nodule | 0.980 | 1.000 | 0.963 | 0.981 | 0.991 |
| Lung opacity | 0.960 | 0.945 | 0.945 | 0.945 | 0.993 |
| Pulmonary fibrotic sequela | 0.960 | 0.918 | 0.957 | 0.938 | 0.986 |
| Pleural effusion | 0.993 | 0.950 | 1.000 | 0.974 | 0.998 |
| Mosaic attenuation pattern | 1.000 | 1.000 | 1.000 | 1.000 | 1.000 |
| Peribronchial thickening | 0.993 | 1.000 | 0.952 | 0.976 | 0.995 |
| Consolidation | 0.973 | 0.885 | 0.958 | 0.920 | 0.992 |
| Bronchiectasis | 0.973 | 0.833 | 1.000 | 0.909 | 0.996 |
| Interlobular septal thickening | 1.000 | 1.000 | 1.000 | 1.000 | 1.000 |

**Supplementary Figure 1.** Machine translation prompts for CT-RATE-JPN creation. The system prompt used in the study, instructing GPT-4o mini to act as a Japanese radiologist and translate radiology report findings from English to Japanese, with specific output formatting requirements.

**System prompt**
あなたは日本の放射線科医です。以下の英語で書かれた放射線レポートの所見欄を和訳してください。
出力は必ず最終行以降に「和訳：n」から始まるようにしてください。

**User prompt**
[Input Radiology Report]

**Supplementary Figure 2.** English translation of the machine translation prompts. The English version of the prompts shown in Supplementary Figure 1, provided to ensure reproducibility of the translation process.

**System prompt**
You are a Japanese radiologist. Please translate the findings section of the following radiology report from English to Japanese.
Your output must begin with 'Translation:\n' in the final line and below.

**User prompt**
[Input Radiology Report]

**Supplementary Figure 3.** System and user prompts used for GPT-4o structured labeling in Japanese. The system prompt defines the AI's role in extracting structured labels from Japanese radiology reports, with detailed instructions for binary classification (0/1) of 18 specific CT findings and output format specifications. The user prompt requests structured label extraction from an input radiology report.

### System prompt

あなたは日本語の読影レポートから構造化ラベルを抽出する専門的なAIです。以下の手順に従って、読影レポートに基づくラベルを生成してください：

1. 各項目について、該当する記述が読影レポート内に明示的または暗示的に存在する場合、その項目に「1」を割り当ててください。
2. 該当する記述がない場合、その項目に「0」を割り当ててください。
3. 結果をJSON形式で出力してください。
4. 出力は項目名（英語）をキーとし、それに対応する値を「0」または「1」で表してください。

対象項目：
- Medical material
- Arterial wall calcification
- Cardiomegaly
- Pericardial effusion
- Coronary artery wall calcification
- Hiatal hernia
- Lymphadenopathy
- Emphysema
- Atelectasis
- Lung nodule
- Lung opacity
- Pulmonary fibrotic sequela
- Pleural effusion
- Mosaic attenuation pattern
- Peribronchial thickening
- Consolidation
- Bronchiectasis
- Interlobular septal thickening

出力例：
```json
{
"Medical material": 1,
"Arterial wall calcification": 1,
"Cardiomegaly": 0,
...
}
```

### User prompt

以下の日本語の読影レポートに基づいて、構造化ラベルを抽出してください：
[Input Radiology Report]

**Supplementary Figure 4.** English translation of the system and user prompts used for GPT-4o structured labeling. This figure shows the complete English version of the prompts presented in Supplementary Figure 3, including the system prompt detailing the AI's role and instructions for binary classification of 18 CT findings, and the user prompt requesting structured label extraction from an input radiology report.

### System prompt

You are a specialized AI that extracts structured labels from Japanese radiology reports. Please generate labels based on radiology reports according to the following steps:

1. Assign "1" to items where relevant descriptions exist explicitly or implicitly in the radiology report.
2. Assign "0" to items where no relevant descriptions exist.
3. Output the results in JSON format.
4. The output should use item names (in English) as keys with corresponding values of "0" or "1".

Target items:
- Medical material
- Arterial wall calcification
- Cardiomegaly
- Pericardial effusion
- Coronary artery wall calcification
- Hiatal hernia
- Lymphadenopathy
- Emphysema
- Atelectasis
- Lung nodule
- Lung opacity
- Pulmonary fibrotic sequela
- Pleural effusion
- Mosaic attenuation pattern
- Peribronchial thickening
- Consolidation
- Bronchiectasis
- Interlobular septal thickening

Output example:
```json
{
"Medical material": 1,
"Arterial wall calcification": 1,
"Cardiomegaly": 0,
...
}
```

### User prompt

Please extract structured labels based on the following Japanese radiology report:
[Input Radiology Report]